\documentclass[10pt,twocolumn,letterpaper]{article}

\usepackage[pagenumbers]{cvpr} 

\usepackage{makecell}
\usepackage{graphicx}
\usepackage{amsmath}
\usepackage{amssymb}
\usepackage{wrapfig, lipsum, booktabs}

\usepackage{multirow}
\usepackage{xcolor}

\usepackage{arydshln}

\usepackage{comment}
\usepackage{listings}
\usepackage{float}
\usepackage{adjustbox}
\usepackage{caption}
\usepackage{color}
\usepackage[ruled,vlined]{algorithm2e}
\usepackage{anyfontsize}

\usepackage[pagebackref,breaklinks,colorlinks]{hyperref}
\hypersetup{
    colorlinks=true,
    urlcolor=cyan,
}
\usepackage[accsupp]{axessibility}

\usepackage[capitalize]{cleveref}
\crefname{section}{Sec.}{Secs.}
\Crefname{section}{Section}{Sections}
\Crefname{table}{Table}{Tables}
\crefname{table}{Tab.}{Tabs.}

\definecolor{commentcolor}{RGB}{110,154,155}   
\definecolor{predefcolor}{RGB}{155,110,110}
\definecolor{classcolor}{RGB}{204, 120, 50}
\definecolor{highlight}{RGB}{150, 0, 0}
\newcommand{\PyComment}[1]{\ttfamily\textcolor{commentcolor}{\# #1}}

\newcommand{\PyDef}[1]{\ttfamily\textcolor{predefcolor}{\# #1}}
\newcommand{\PyNline}[1]{\ttfamily\textcolor{commentcolor}{#1}}
\newcommand{\PyCode}[1]{\ttfamily\textcolor{black}{#1}}
\newcommand{\PyClass}[1]{\ttfamily\textcolor{orange}{#1}}

\begin{document}

\title{itKD: Interchange Transfer-based Knowledge Distillation for 3D Object Detection}

\author{Hyeon Cho$^1$, Junyong Choi$^{1,2}$, Geonwoo Baek$^1$, and Wonjun Hwang$^{1,3}$\\
$^1$Ajou University, $^2$Hyundai Motor Company, $^3$Naver AI Lab\\
{\tt\small ch0104@ajou.ac.kr, chldusxkr@hyundai.com, bkw0622@ajou.ac.kr, wjhwang@ajou.ac.kr}
}
\maketitle

\begin{abstract}
Point-cloud based 3D object detectors recently have achieved remarkable progress. However, most studies are limited to the development of network architectures for improving only their accuracy without consideration of the computational efficiency. In this paper, we first propose an autoencoder-style framework comprising channel-wise compression and decompression via interchange transfer-based knowledge distillation. To learn the map-view feature of a teacher network, the features from teacher and student networks are independently passed through the shared autoencoder; here, we use a compressed representation loss that binds the channel-wised compression knowledge from both student and teacher networks as a kind of regularization. The decompressed features are transferred in opposite directions to reduce the gap in the interchange reconstructions. Lastly, we present an head attention loss to match the 3D object detection information drawn by the multi-head self-attention mechanism. Through extensive experiments, we verify that our method can train the lightweight model that is well-aligned with the 3D point cloud detection task and we demonstrate its superiority using the well-known public datasets; e.g., Waymo and nuScenes.\footnote{Our code is available at \href{https://github.com/hyeon-jo/interchange-transfer-KD}{https://github.com/hyeon-jo/interchange-transfer-KD.}}
\end{abstract}

\section{Introduction}
Convolutional neural network (CNN)-based 3D object detection methods using point clouds~\cite{PP}\cite{SECOND}\cite{PIXOR}\cite{CPoint}\cite{VoxelNet} have attracted wide attention based on their outstanding performance for self-driving cars. Recent CNN-based works have required more computational complexity to achieve higher precision under the various wild situation. Some studies~\cite{PointRCNN}\cite{PIXOR}\cite{CPoint} have proposed methods to improve the speed of 3D object detection through which the non-maximum suppression (NMS) or anchor procedures are removed but the network parameters are still large.

\begin{figure}[t]
\centering
\includegraphics[width=0.9\linewidth]{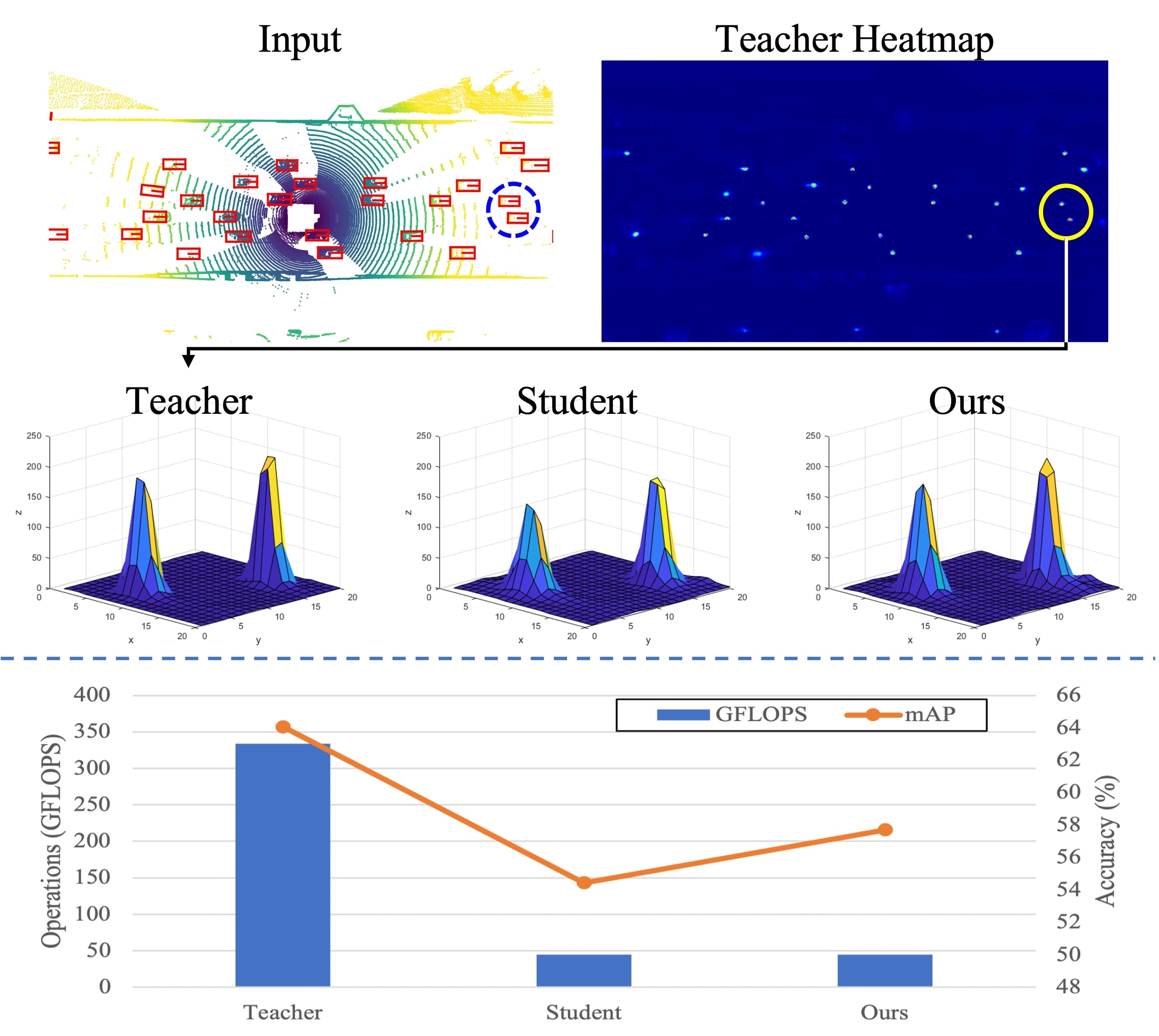}
\vspace{-0.2cm}
\caption{\textbf{Performance comparison between teacher and student networks for a point-cloud based 3D object detection.} The top example images are qualitatively compared between the results of teacher, student and our networks. Specifically, the first row images are an input sample with labels and the center heatmap head of the teacher network. The second row examples are responses of teacher, student, and ours for the yellow circle on the heatmap (or the blue dash circle on the input). The bottom image quantitatively shows the computational complexity and the corresponding accuracy of teacher, student and our networks, respectively. Best viewed in color.
}
\vspace{-0.4cm}
\label{fig:fig1}
\end{figure}

Knowledge distillation (KD) is one of the parameter compression techniques, which can effectively train a compact student network through the guidance of a deep teacher network, as shown in the example images of Fig.~\ref{fig:fig1}.
Starting with Hinton's work~\cite{hintonKD}, many KD studies~\cite{IEKD}\cite{FitNet}\cite{DGKD}\cite{AttentionTransfer} have transferred the discriminative teacher knowledge to the student network for classification tasks. From the viewpoint of the detection task, KD should be extended to the regression problem, including the object locations, which is not easy to straight-forwardly apply the classification-based KD methods to the detection task. To alleviate this problem, KD methods for object detection have been developed for mimicking the output of the backbone network~\cite{MimicKD} (\textit{e.g}., region proposal network) or individual detection head~\cite{detKD}\cite{wang2019distilling}. Nevertheless, these methods have only been studied for detecting 2D image-based objects, and there is a limit to applying them to sparse 3D point cloud-based data that have not object-specific color information but only 3D position-based object structure information. 

Taking a closer look at differences between 2D and 3D data, there is a large gap in that 2D object detection usually predicts 2D object locations based on inherent color information with the corresponding appearances, but 3D object detection estimates 3D object boxes from inputs consisting of only 3D point clouds. Moreover, the number of the point clouds constituting objects varies depending on the distances and presence of occlusions~\cite{yihan2021learning}.
Another challenge in 3D object detection for KD is that, compared to 2D object detection, 3D object detection methods~\cite{dai2021dynamic}\cite{ge2020afdet}\cite{CPoint}\cite{rukhovich2022imvoxelnet} have more detection head components such as 3D boxes, and orientations. These detection heads are highly correlated with each other and represent different 3D characteristics. In this respect, when transferring the detection heads of the teacher network to the student network using KD, it is required to guide the distilled knowledge under the consideration of the correlation among the multiple detection head components. 

In this paper, we propose a novel interchange transfer-based KD (itKD) method designed for the lightweight point-cloud based 3D object detection. The proposed itKD comprises two modules: (1) a channel-wise autoencoder based on the interchange transfer of reconstructed knowledge and (2) a head relation-aware self-attention on multiple 3D detection heads. First of all, through a channel-wise compressing and decompressing processes for KD, the interchange transfer-based autoencoder effectively represents the map-view features from the viewpoint of 3D representation centric-knowledge. Specifically, the encoder provides an efficient representation by compressing the map-view feature in the channel direction to preserve the spatial positions of the objects and the learning of the student network could be regularized by the distilled position information of objects in the teacher network. For transferring the interchange knowledge to the opposite networks, the decoder of the student network reconstructs the map-view feature under the guidance of the teacher network while the reconstruction of the teacher network is guided by the map-view feature of the student network. As a result, the student network can effectively learn how to represent the 3D map-view feature of the teacher.
Furthermore, to refine the teacher’s object detection results as well as its representation, our proposed head relation-aware self-attention gives a chance to learn the pivotal information that should be taught to the student network for improving the 3D detection results by considering the inter-head relation among the multiple detection head and the intra-head relation of the individual detection head.

In this way, we implement a unified KD framework to successfully learn the 3D representation and 3D detection results of the teacher network for the lightweight 3D point cloud object detection. We also conduct extensive ablation studies for thoroughly validating our approach in Waymo and nuScenes datasets. The results reveal the outstanding potential of our approach for transferring distilled knowledge that can be utilized to improve the performance of 3D point cloud object detection models.

Our contributions are summarized as follows:
\begin{itemize}
    \setlength{\itemsep}{0pt}%
    \setlength{\parskip}{0pt}%
    \vspace{-.5em}
    \item For learning the 3D representation-centric knowledge from the teacher network, we propose the channel-wise autoencoder regularized in the compressed domain and the interchange knowledge transfer method wherein the reconstructed features are guided by the opposite networks. 
    \item For detection head-centric knowledge of the teacher, we suggest the head relation-aware self-attention which can efficiently distill the detection properties under the consideration of the inter-head relation and intra-head relation of the multiple 3D detection heads.
    \item Our work is the best attempt to reduce the parameters of point cloud-based 3D object detection using KD. Additionally, we validate its superiority using two large datasets that reflect real-world driving conditions, e.g., Waymo and NuScenes.
\end{itemize}

\section{Related Works}
\subsection{3D Object Detection based on Point Cloud}

During the last few years, encouraged by the success of CNNs, the development of object detectors using CNNs is developing rapidly. Recently, many 3D object detectors have been studied and they can be briefly categorized by how they extract representations from point clouds; \textit{e.g.}, 
grid-based~\cite{SECOND}\cite{PIXOR}\cite{VoxelNet}\cite{PP}\cite{CPoint},
point-based~\cite{FPointNet}\cite{PointRCNN}\cite{StarNet}\cite{PointGNN}\cite{3DSSD}
and hybrid-based~\cite{FPRCNN}\cite{STD}\cite{SASSD}\cite{MVF}\cite{PVRCNN}
methods. 
In detail, Vote3Deep~\cite{Vote3Deep} thoroughly exploited feature-centric voting to build CNNs for detecting objects in point clouds. 
In~\cite{DSS}, they have studied on the task of amodal 3D object detection in RGB-D images, where a 3D region proposal network (RPN) to learn objectness from geometric shapes and the joint object recognition network to extract geometric features in 3D and color features in 2D. 
The 3D fully convolutional network~\cite{3DFCN} was straightforwardly applied to point cloud data for vehicle detection.
In the early days, VoxelNet~\cite{VoxelNet} has designed an end-to-end trainable detector based on learning-based voxelization using fully connected layers.
In~\cite{SECOND}, they encoded the point cloud by VoxelNet and used the sparse convolution to achieve the fast detection. 
HVNet~\cite{HVNet} fused the multi-scale voxel feature encoder at the point-wise level and projected into multiple pseudo-image feature maps for solving the various sizes of the feature map. 
In~\cite{ComplexYOLO}, they replaced the point cloud with a grid-based bird’s-eye view (BEV) RGB-map and utilized YOLOv2 to detect the 3D objects.
PIXOR~\cite{PIXOR} converted the point cloud to a 3D BEV map and carried out the real-time 3D object detection with an RPN-free single-stage based model.

Recently, PointPillars (PP)-based method~\cite{PP} utilized the PointNet~\cite{PointNet} to learn the representation of point clouds organized in vertical columns for achieving the fast 3D object detection. To boost both performance and speed over PP, a pillar-based method~\cite{PBOD} that incorporated a cylindrical projection into multi-view feature learning was proposed.
More recently, CenterPoint~\cite{CPoint} was introduced as an anchor-free detector that predicted the center of an object using a PP or VoxelNet-based feature encoder. 
In this paper, we employ the backbone architecture using CenterPoint because it is simple, near real-time, and achieves good performance in the wild situation.

\subsection{Knowledge Distillation}

KD is one of the methods used for compressing deep neural networks and its fundamental key is to imitate the knowledge extracted from the teacher network, which has heavy parameters as well as good accuracy. 
Hinton et al.~\cite{hintonKD} performed a knowledge transfer using KL divergence; FitNet~\cite{FitNet} proposed a method for teaching student networks by imitating intermediate layers.
On the other hand, TAKD~\cite{Mirzadeh20} and DGKD~\cite{DGKD} used multiple teacher networks for transferring more knowledge to the student network in spite of large parameter gaps.
Recently, some studies have been proposed using the layers shared between the teacher and the student networks for KD.
Specifically, in~\cite{softmaxReg}, KD was performed through softmax regression as the student and teacher networks shared the same classifier.
IEKD~\cite{IEKD} proposed a method to split the student network into inheritance and exploration parts and mimic the compact teacher knowledge through a shared latent feature space via an autoencoder. 

Beyond its use in classification, KD for detection should transfer the regression knowledge regarding the positions of the objects to the student network. For this purpose, a KD for 2D object detection~\cite{MimicKD} was first proposed using feature map mimic learning. In~\cite{detKD}, they transferred the detection knowledge of the teacher network using hint learning for an RPN, weighted cross-entropy loss for classification, and bound regression loss for regression. Recently, Wang et al.~\cite{wang2019distilling} proposed a KD framework for detection by utilizing the cross-location discrepancy of feature responses through fine-grained feature imitation. 

As far as we know, there are few KD studies~\cite{LIGA}\cite{zheng2021se}\cite{objDGCNN}\cite{yang2022towards} on point cloud-based 3D object detection so far. 
However, looking at similar studies on 3D knowledge transfer, 
SE-SSD~\cite{zheng2021se} presented a knowledge distillation-based self-ensembling method for exploiting soft and hard targets with constraints to jointly optimize the model without extra computational cost during inference time.
Object-DGCNN~\cite{objDGCNN} proposed a NMS-free 3D object detection via dynamic graphs and a set-to-set distillation. They used the set-to-set distillation method for improving the performance without the consideration of the model compression. Another latest study is SparseKD~\cite{yang2022towards} which suggested a label KD method that distills a few pivotal positions determined by teacher classification response to enhance the logit KD method. 
On the other hand, in this paper, we are more interest in how to make the student network lighter, or lower computational complexity, by using the KD for 3D object detection.

\begin{figure*}[t]
\centering
\includegraphics[width=0.8\linewidth]{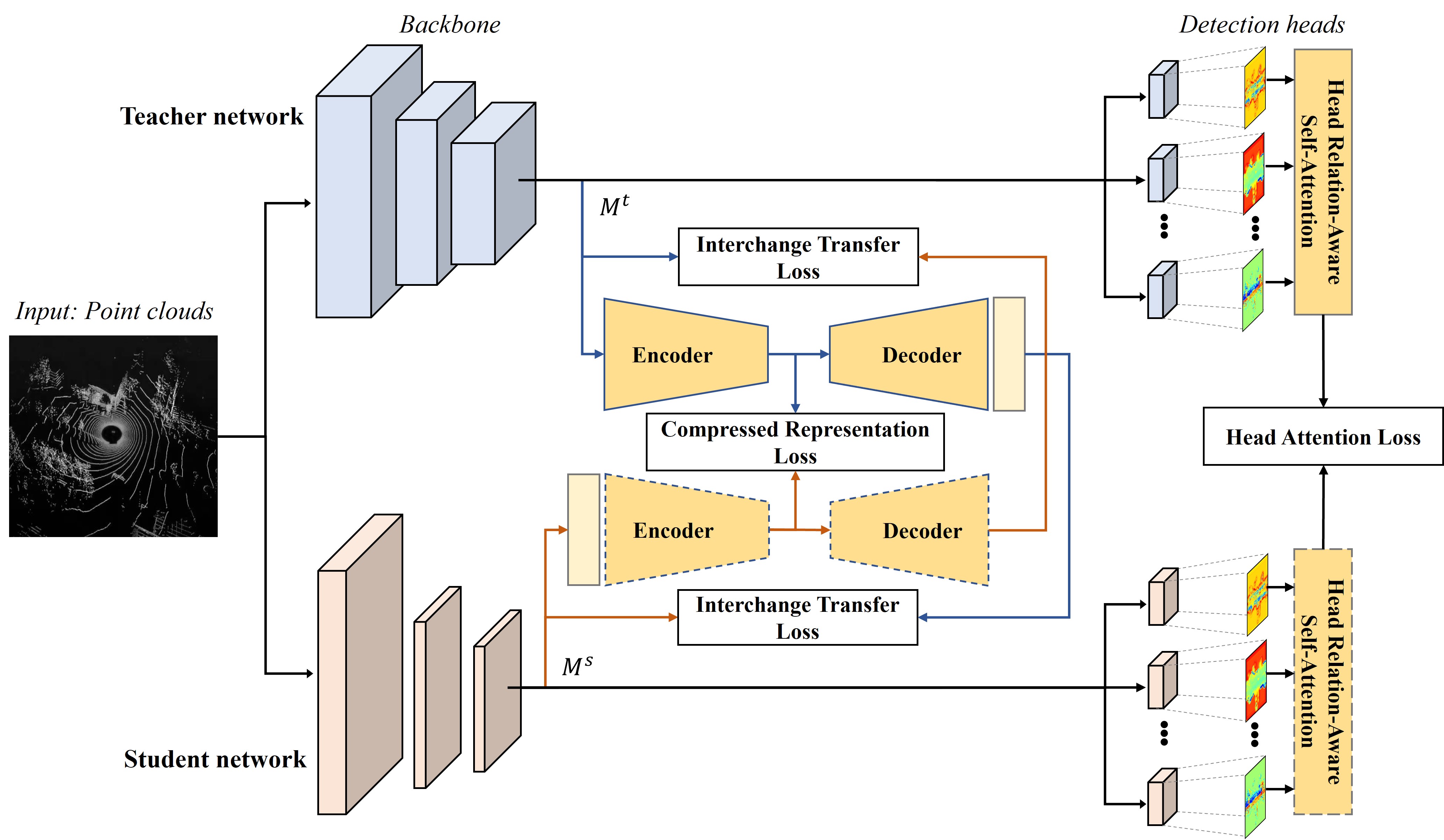}
\vspace{-0.2cm}
\caption{\textbf{Overview of the proposed knowledge distillation method.} 
The teacher and student networks take the same point clouds as inputs. Then, the map-view features $M^t$ and $M^s$ are extracted from the teacher and student networks, respectively. 
The channel-wise autoencoder transfers the knowledge obtained from $M^t$ to $M^s$ by using the compressed representation loss and interchange transfer loss consecutively. The head relation-aware self-attention provides the relation-aware knowledge of multiple detection head to the student network using the attention head loss.
The dotted lines of the modules denote that there are shared network parameters between the teacher and student networks. The light-yellow boxes are buffer layers for sampling the features to match the channel sizes of networks.
}
\vspace{-0.4cm}
\label{fig:overall}
\end{figure*}

\section{Methodology} \label{sec3}
\subsection{Background}
The 3D point cloud object detection methods~\cite{PP}\cite{VoxelNet} generally consists of three components; a point cloud encoder, a backbone network, and detection heads. In this paper, we employ CenterPoint~\cite{CPoint} network as a backbone architecture. Since the parameter size of the backbone network\footnote{The total parameter size of the 3D detector is about 5.2M and the backbone size is approximately 4.8M, which is 92\%. Further details are found in the supplementary material.} is the largest among components of the 3D object detector, 
we aim to construct the student network by reducing the channel sizes of the backbone network for efficient network. 
We design our method to teach the student 3D representation-centric knowledge and detection head-centric knowledge of the teacher network, respectively.

\subsection{Interchange Transfer}
We adopt an autoencoder framework to effectively transfer the meaningful distilled knowledge regarding 3D detection from the teacher to the student network. 
The traditional encoder-based KD methods~\cite{IEKD}\cite{kim2018paraphrasing} have been limited to the classification task, which transfers only compressed categorical knowledge to the student network. 
However, from the viewpoint of the detection task, the main KD goal of this paper is transferring the distilled knowledge regarding not only categorical features but also object location-related features. 
Particularly, unlike 2D detectors, 3D object detectors should regress more location information such as object orientations, 3D box sizes, etc., and it results in increasing the importance of how to transfer the 3D location features to the student network successfully.

For this purpose, we transfer the backbone knowledge that contains 3D object representation from the teacher network to the student through the compressed and reconstructed knowledge domains.
As shown in Fig.~\ref{fig:overall}, we introduce a channel-wise autoencoder
which consists of an encoder in which the channel dimension of the autoencoder is gradually decreased and a decoder in the form of increasing the channel dimension. Note that spatial features play a pivotal role in the detection task and we try to preserve the spatial information by encoding features in the channel direction.
We propose a compressed representation loss to coarsely guide location information of the objects to the student network in Fig.~\ref{fig:overall}, and the compressed representation loss has an effect similar to the regularization of the autoencoder that binds the coordinates of the objectness between the teacher and student networks.
The compressed representation loss function $\mathcal{L}_{cr}$ is represented as follows:
\begin{equation}\label{enc_loss}
\begin{split}
\mathcal{L}_{cr}&=m_{obj}\circ\mathcal{S}[E(\theta_{enc}, M^{t}), E(\theta_{enc}, M^{s})] \\
&=m_{obj}\circ\mathcal{S}[M_{enc}^{t}, M_{enc}^{s}], 
\end{split}
\end{equation}
where $E$ is a shared encoder, which has the parameters $\theta_{enc}$, and $\mathcal{S}$ denotes $l_{1}$ loss as a similarity measure. $M^t$ and $M^s$ are outputs of the teacher and student backbones, respectively. $m_{obj}$ represents a binary mask to indicate object locations in backbone output like~\cite{yang2022towards} and $\circ$ is an element-wise product.

\begin{figure*}[t]
\centering
\includegraphics[width=0.8\linewidth]{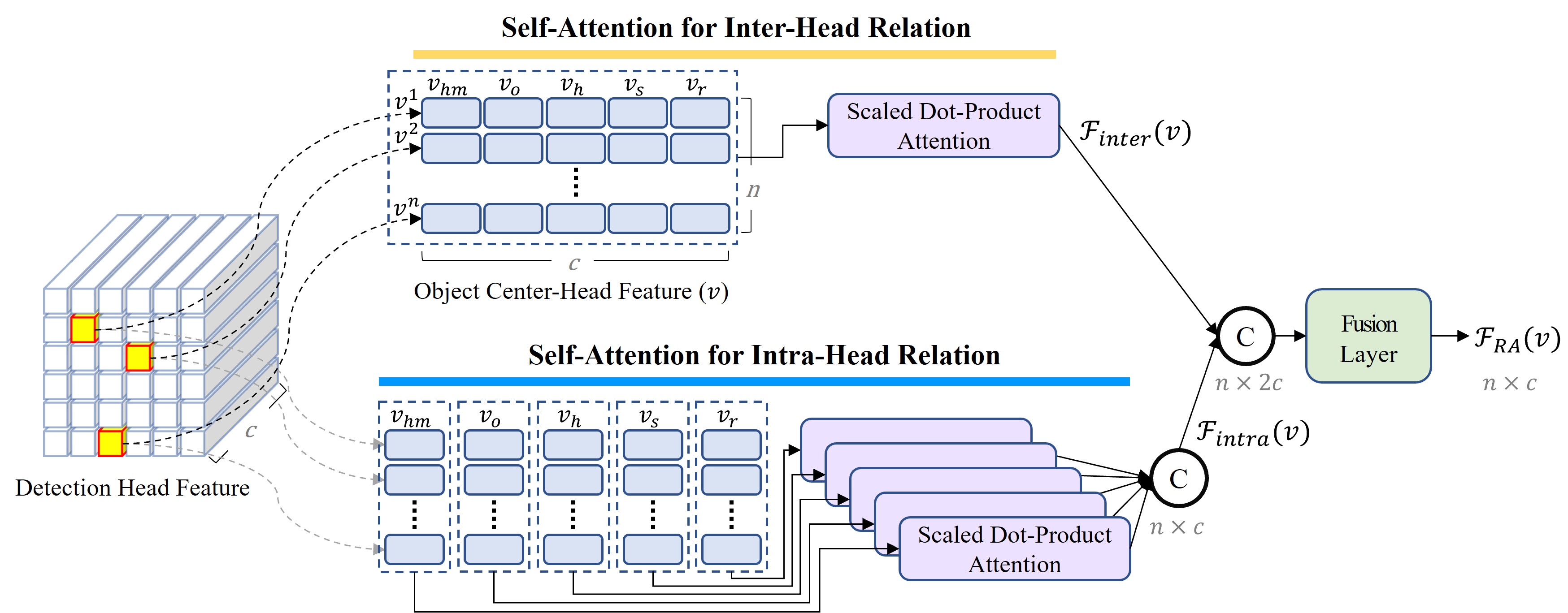} 
\vspace{-0.2cm}
\caption{\textbf{Head Relation-Aware Self-Attention.} We make the object center-head feature from object center locations in the detection head feature and use it as different shaped inputs to self-attentions for inter-head relation and intra-head relation. In the self-attention for inter-head relation, we use the object center-head feature as an input for the self-attention. In the self-attention for intra-head relation, the detection heads are separately used for the independent self-attention functions. The outputs of the self-attentions are concatenated by \textcircled{${c}$} operations and the head relation-aware self-attention is generated through the fusion layer. 
}
\vspace{-0.4cm}
\label{fig:chattn}
\end{figure*}

After performing the coarse representation-based knowledge distillation in a compressed domain, the fine representation features of the teacher network are required to teach the student network from the viewpoint of 3D object detection. In this respect, the decoder reconstructs the fine map-view features in the channel direction from the compressed features. Through the proposed interchange transfer loss, the reconstructed features are guided from the opposite networks, not their own stem networks, as shown in Fig.~\ref{fig:overall}. 
Specifically, since the teacher network is frozen and we use the shared autoencoder for both student and teacher networks, we can teach the reconstructed fine features from the student network to resemble the output of the teacher network $M^t$ rather than the student $M^s$. Moreover, the reconstructed fine features from the teacher network can guide the student's output, $M^s$ at the same time.
The proposed interchange transfer loss $\mathcal{L}_{it}$ is defined as follows:
\begin{equation} \label{t2s_loss}
    \mathcal{L}_{t\to s} = \mathcal{S}[M^{s}, D(\theta_{dec}, M^{t}_{enc})],
\end{equation}
\vspace{-0.5cm}
\begin{equation} \label{s2t_loss}
    \mathcal{L}_{s\to t} = \mathcal{S}[M^{t}, D(\theta_{dec}, M^{s}_{enc})],
\end{equation}
\vspace{-0.5cm}
\begin{equation} \label{icr_loss}
    \mathcal{L}_{it} = \mathcal{L}_{s\to t} + \mathcal{L}_{t\to s},
\end{equation}
where $D$ is the decoder that contains the network parameter $\theta_{dec}$, which is a shared parameter. 
We hereby present the representation-based KD for 3D object detection in both compressed and decompressed domains to guide the student network to learn the map-view feature of the teacher network efficiently. 

\subsection{Head Relation-Aware Self-Attention}

Fundamentally, our backbone network, \textit{e.g.}, CenterPoint~\cite{CPoint}, has various types of 3D object characteristics on detection heads. Specifically, the locations, size, and  direction of an object are different properties, but they are inevitably correlated to each other because they come from the same object. 
However, the traditional KD methods~\cite{detKD}\cite{objDGCNN} were only concerned with how the student network straight-forwardly mimicked the outputs of the teacher network without considering the relation among the detection heads. To overcome this problem, we make use of the relation of detection heads as a major factor for the detection head-centric KD.  

Our proposed head relation-aware self-attention is directly inspired by the multi-head self-attention~\cite{transformer} in order to learn the relation between the multiple detection head. As shown in Fig.~\ref{fig:chattn}, we first extract $i$-th instance feature $v^i \in \mathbb{R}^{c}$, where $c$ is the channel size, from the center location of the object in the detection head feature. Note that, since the instance feature is extracted from the multiple detection head, it has several object properties such as a class-specific heatmap $v^i_{hm}$, a sub-voxel location refinement $v^i_o$, a height-above-ground $v^i_h$, a 3D size $v^i_s$, and a yaw rotation angle $v^i_r$. When there are a total of $n$ objects, we combine them to make an object center-head feature $v \in \mathbb{R}^{n\times c}$. 
We use the same object center-head feature $v$ of dimension $n$ for query, key, and value, which are an input of the scaled dot-product attention. The self-attention function $\mathcal{F}$ is computed by 
\begin{equation}\label{self_attn}
    \mathcal{F}(v) = softmax(\frac{v^{\top} \cdot v}{{\sqrt{n}}})\cdot v.
\end{equation}

The proposed head relation-aware self-attention consists of two different self-attentions for inter-head and intra-head relations as illustrated in Fig.~\ref{fig:chattn}.
We propose the self-attention based on the inter-head relation of the instance features, which is made in order to consider the relation between all detected objects and their different properties, rather than a single detected instance, from the global viewpoint. 
The self-attention for inter-head relation is computed by
\begin{equation}\label{intra_attn}
    \mathcal{F}_{inter}(v) = \mathcal{F}([v_{hm}, v_{o}, v_{h}, v_{s}, v_{r}]).
\end{equation}

On the other hand, we suggest the self-attention for intra-head relation using the individual detection heads. Here we perform the attentions using only local relation in individual detection heads designed for different properties (e.g., orientation, size, etc.) and concatenate them. Its equation is 
\begin{equation}\label{inter_attn}
    \mathcal{F}_{intra}(v) = [\mathcal{F}(v_{hm}), \mathcal{F}(v_{o}), \mathcal{F}(v_{h}), \mathcal{F}(v_{s}), \mathcal{F}(v_{r})].
\end{equation}

We concatenate the outputs of the self-attentions and apply the fusion layer to calculate a final attention score that considers the relation between the detection heads and objects. The head relation-aware self-attention equation $\mathcal{F}_{RA}$ is derived by:
\begin{equation}\label{ch_attn}
    \mathcal{F}_{RA}(v) = \mathcal{G}([\mathcal{F}_{inter}(v), \mathcal{F}_{intra}(v)]),
\end{equation}
where 
$\mathcal{G}$ is the fusion layer, e.g., 1$\times$1 convolution layer.
The student network indirectly takes the teacher's knowledge by learning the relation between the multiple detection head of the teacher network through head attention loss as follows:
\begin{equation}\label{attn_distill}
    \mathcal{L}_{attn}=\mathcal{S}(\mathcal{F}_{RA}(v_t), \mathcal{F}_{RA}(v_s)),
\end{equation}
where $v_t$ and $v_s$ are the object center-head features of the teacher and the student, respectively. 

Consequently, the overall loss is derived by
\begin{equation}
\mathcal{L}_{total} = \alpha \mathcal{L}_{sup} + \beta (\mathcal{L}_{it} + \mathcal{L}_{cr} + \mathcal{L}_{attn}),
\label{eq:total_loss}
\end{equation}
where $\mathcal{L}_{sup}$ is the supervised loss that consists of focal loss and regression loss, and $\alpha$ and $\beta$ are the balancing parameters, which we set as 1 for simplicity.  

\section{Experimental Results and Discussions}
\subsection{Environment Settings}\label{sec4}
\textbf{Waymo} Waymo open dataset~\cite{sun2020scalability} is one of the large-scale datasets for autonomous driving, which is captured by the synchronized and calibrated high-quality LiDAR and camera across a range of urban and suburban geographies. This dataset provides 798 training scenes and 202 validation scenes obtained by detecting all the objects within a 75m radius; it has a total of 3 object categories (\textit{e.g.}, vehicle, pedestrian, and cyclist) which have 6.1M, 2.8M, and 67K sets, respectively. 
The mean Average Precision (mAP) and mAP weighted by heading accuracy (mAPH) are the official metrics for Waymo evaluation. mAPH is a metric that gives more weight to the heading than it does to the sizes, and it accounts for the direction of the object.

\textbf{nuScenes} nuScenes dataset~\cite{caesar2020nuscenes} is another large-scale dataset used for autonomous driving. This dataset contains 1,000 driving sequences. 700, 150, and 150 sequences are used for training, validation, and testing, respectively. Each sequence is captured approximately 20 seconds with 20 FPS using the 32-lane LiDAR. Its evaluation metrics are the average precision (AP) and nuScenes detection score (NDS). NDS is a weighted average of mAP and true positive metrics which measures the quality of the detections in terms of box location, size, orientation, attributes, and velocity.

\textbf{Implementation details} Following the pillar-based CenterPoint~\cite{CPoint} as the teacher network, we use an Adam optimizer~\cite{kingma2014adam} with a weight decay of 0.01 and a cosine annealing strategy~\cite{smith2017cyclical} to adjust the learning rate. We set 0.0003 for initial learning rate, 0.003 for max learning rate, and 0.95 for momentum. The networks have been trained for 36 epochs on 8$\times$V100 GPUs with a batch size of 32. 
For Waymo dataset, we set the detection range to [-74.88m, 74.88m] for the X and Y axes, [-2m, 4m] for the Z-axis, and a grid size of (0.32m, 0.32m). In experiments on nuScenes dataset, we used a (0.2m, 0.2m) grid and set the detection range to [-51.2m, 51.2m] for the X and Y-axes, [-5m, 3m] for the Z-axis, and a grid size of (0.2m, 0.2m).
Compared to the teacher network, the student network has ${1}/{4}$ less channel capacity of backbone network.
Our channel-wise autoencoder consists of three 1$\times$1 convolution layers as the encoder and three 1$\times$1 convolution layers as the decoder and the number of filters are {128, 64, 32} in encoder layers and {64, 128, 384} in decoder layers. The student's input buffer layer increases the channel size of 196 to 384 and the teacher's output buffer layer decreases the channel size 384 to 196.

\subsection{Overall KD Performance Comparison}

\begin{table*}[t]
\caption{\textbf{Waymo evaluation.} Comparisons with different KD methods in the Waymo validation set. The best accuracy is indicated in bold, and the second-best accuracy is underlined.} 
\label{table:Waymo_sota}
\vspace{-3mm}
\resizebox{\textwidth}{!}{
    \begin{tabular}{c|cccc:cccc:cccc}
\specialrule{.1em}{.05em}{.05em} 
\multirow{3}{*}{Method} & \multicolumn{4}{c:}{Vehicle}                               & \multicolumn{4}{c:}{Pedestrian}                            & \multicolumn{4}{c}{Cyclist}                               \\
                        & \multicolumn{2}{c}{Level 1} & \multicolumn{2}{c:}{Level 2} & \multicolumn{2}{c}{Level 1} & \multicolumn{2}{c:}{Level 2} & \multicolumn{2}{c}{Level 1} & \multicolumn{2}{c}{Level 2} \\
                            & mAP          & mAPH         & mAP          & mAPH        & mAP          & mAPH         & mAP          & mAPH         & mAP          & mAPH         & mAP          & mAPH      \\ 
                            \hline\hline
Teacher~\cite{CPoint}       & 73.72        & 73.17        & 65.61        & 65.11       & 72.43        & 61.72        & 64.73        & 54.99        & 64.30        & 62.61        & 61.91       & 60.28      \\
Student (${1} / {4}$)                    & 64.22        & 63.56        & 56.21        & 55.62       & 63.72        & 53.22        & 56.14        & 46.78        & 53.01        & 51.72        & 50.99       & 49.75      \\ \hline
Baseline    & 64.78        & 64.05        & 56.92        & 56.26       & 64.85        & 52.98        & 57.37        & 46.75        & 54.71        & 52.46        & 52.65       & 50.48      \\
FitNet~\cite{FitNet}        & 65.11        & 64.38        & 57.24        & 56.58       & 64.89        & 53.29        & 57.37        & 47.00        & 54.91        & 52.61        & 52.84       & 50.63      \\
EOD-KD~\cite{detKD}         & \underline{66.50}        & \underline{65.79}        & 58.56        & 57.92       & 65.99        & 54.58        & 58.48        & 48.25        & 55.18        & 52.93        & 53.10       & 50.94      \\
SE-SSD~\cite{zheng2021se} & 65.95 & 65.22 & 58.05 & 57.40 & 65.39 & 53.98   & 57.92           & 47.69      & 55.01     & 52.98   & 52.94 & 50.99   \\
TOFD~\cite{TOFD}            & 64.09        & 63.43       & 56.13        &  55.55      & 66.24        & \underline{54.98}        & 58.50        &    48.45     & 54.95        & 53.06        & 52.86       & 51.04      \\
Obj. DGCNN~\cite{objDGCNN}   & 66.07        & {65.38}        & \underline{59.27}        & \underline{58.55}       & 65.98        & 54.44        & \underline{59.42}        & \underline{49.11}        & 54.65        & 52.62        & 53.13       & 50.93      \\
 SparseKD~\cite{yang2022towards} & 65.25 & 64.59 & 56.97 & 56.38 & \textbf{67.44} & 54.54   & 59.24           & 47.83      & \underline{55.54}     & \underline{53.45}   & \underline{53.63} & \underline{51.61}   \\
\hline
Ours                        & \textbf{67.43}        & \textbf{66.72}        & \textbf{59.44}        & \textbf{58.81}       & \underline{67.26}        & \textbf{56.02}        & \textbf{59.73}        & \textbf{49.61}        & \textbf{56.09}        & \textbf{54.24}        & \textbf{53.96}       & \textbf{52.19}      \\ 
\specialrule{.1em}{.05em}{.05em} 
    \end{tabular}
}
\vspace{-0.1cm}
\end{table*}

\begin{table*}[t]
\caption{\textbf{nuScenes evaluation.} Comparisons with different KD methods in the nuScenes validation set. The best accuracy is indicated in bold, and the second-best accuracy is underlined.} 
\label{table:nuscenes_sota}
\vspace{-3mm}
\resizebox{\textwidth}{!}{
    \begin{tabular}{c|cc|cccccccccc}
\specialrule{.1em}{.05em}{.05em} 
Method                     & NDS   & mAP   & car   & truck & bus   & trailer & con. veh. & ped. & motor. & bicycle & tr. cone  & barrier \\
\hline\hline
Teacher~\cite{CPoint}      & 60.16 & 50.25 & 84.04 & 53.48 & 64.29 & 31.90   & 12.50          & 78.93      & 44.01     & 18.18   & 54.87 & 60.30   \\
Student (${1} / {4}$)  & 50.24 & 38.52 & 77.85 & 38.18 & 51.38 & 22.33   & 3.95           & 71.51      & 23.90     & 3.51    & 43.03 & 49.56   \\ \hline
Baseline   & 51.48 & 39.19 & 78.72 & 37.90 & 50.47 & 22.42   & 3.51           & 72.29      & 26.25     & 4.65   & 44.91 & 50.77   \\
FitNet~\cite{FitNet}       & 51.42 & 38.90 & 78.30 & 37.40 & 50.40 & 22.20   & 3.80           & 72.10      & 25.70     & 4.25   & 44.20 & 50.60   \\
EOD-KD~\cite{detKD}        & 52.49 & 39.82 & 78.40 & 38.60 & 50.90 & 22.70   & \underline{3.90}           & 73.20     & 28.20     & 5.30   & 45.00 & 51.97   \\
SE-SSD~\cite{zheng2021se} & 52.21 & 39.53 & 78.69 & 38.56 & 49.81 & 23.70   & 3.72           & 72.86      & 28.27     & 4.25   & 44.24 & 51.18   \\
TOFD~\cite{TOFD}           & 52.88 & \underline{40.57} & \underline{79.06} & \underline{39.73} & 52.03 & \underline{24.51}   & 3.56           & \underline{73.51}      & \underline{29.58}     & \underline{5.62}   & \underline{45.34} & \underline{52.79}   \\
Obj. DGCNN~\cite{objDGCNN} & 52.91 & 40.34 & 78.95 & 39.24 & \underline{53.37} & 23.96   & \textbf{4.13}           & 72.98      & 28.63     & 4.99   & 44.72 & 52.46   \\
SparseKD~\cite{yang2022towards} & \underline{53.01} & 40.26 & 78.78 & 39.50 & 51.87 & 23.64   & 3.30           & 73.17      & 29.34     & \textbf{5.75}   & 44.98 & 52.26   \\
\hline
Ours                       & \textbf{53.90} & \textbf{41.33} & \textbf{79.48} & \textbf{40.38} & \textbf{54.35} & \textbf{26.44}   & 3.58           & \textbf{73.91}      & \textbf{30.21}     & 5.39   & \textbf{45.90} & \textbf{53.70} \\  
\specialrule{.1em}{.05em}{.05em} 
    \end{tabular}
}
\vspace{-0.4cm}
\end{table*}

We validate the performance of our method compared with well-known KD methods on the Waymo and nuScenes datasets.
We re-implement the seven KD methods from 2D classification-based KD to 3D detection-based KD in this paper. 
We set the baseline by applying the Kullback-Leibler (KL) divergence loss~\cite{hintonKD} to the center heatmap head and $l_1$ loss to the other regression heads.
FitNet~\cite{FitNet} is a method that mimics the intermediate outputs of layers and we apply it to the output of the backbone for simplicity. We also simply extend EOD-KD~\cite{detKD}, one of the 2D object detection KDs, to 3D object detection. We apply TOFD~\cite{TOFD}, a 3D classification-based KD, to our detection task and straight-forwardly use SE-SSD~\cite{zheng2021se}, Object DGCNN~\cite{objDGCNN}, and SparseKD~\cite{yang2022towards} for 3D object detection KD. 

Table~\ref{table:Waymo_sota} shows that our method almost outperforms other KD methods on mAP and mAPH values for level 1 and level 2 under all three categories of objects. Especially, our performance improvement of mAPH is better than other methods, which indicates our method guides the student network well where the detected objects are facing.
To verify the generality of the proposed method, we make additional comparison results using the nuScenes dataset, another large-scale 3D dataset for autonomous driving, in Table~\ref{table:nuscenes_sota}. Compared with the other methods, our method achieves the best accuracy under the NDS and mAP metrics in the nuScenes validation set. Specifically, when the student network shows 50.24\% NDS and 38.52\% mAP, our method achieves 53.90\% (+3.66\%) NDS and 41.33\% (+2.81\%) mAP. 
In detail, our method outperforms the other methods for the most of object classes except the construction vehicle and the bicycle.

\subsection{Ablation Studies}

To analyze of our proposed method in detail, we conduct ablation studies on the Waymo dataset, and the whole performances are measured by mAPH at level 2 for simplicity. 
For the qualitative analysis, we visualize the map-view feature at each stage to validate the what kinds of knowledge are transferred from the teacher to the student by the proposed method. For simple visualization, we apply the $L_1$ normalization to the map-view feature in the channel direction.

\begin{figure}[t]
\centering
\includegraphics[width=0.9\linewidth]{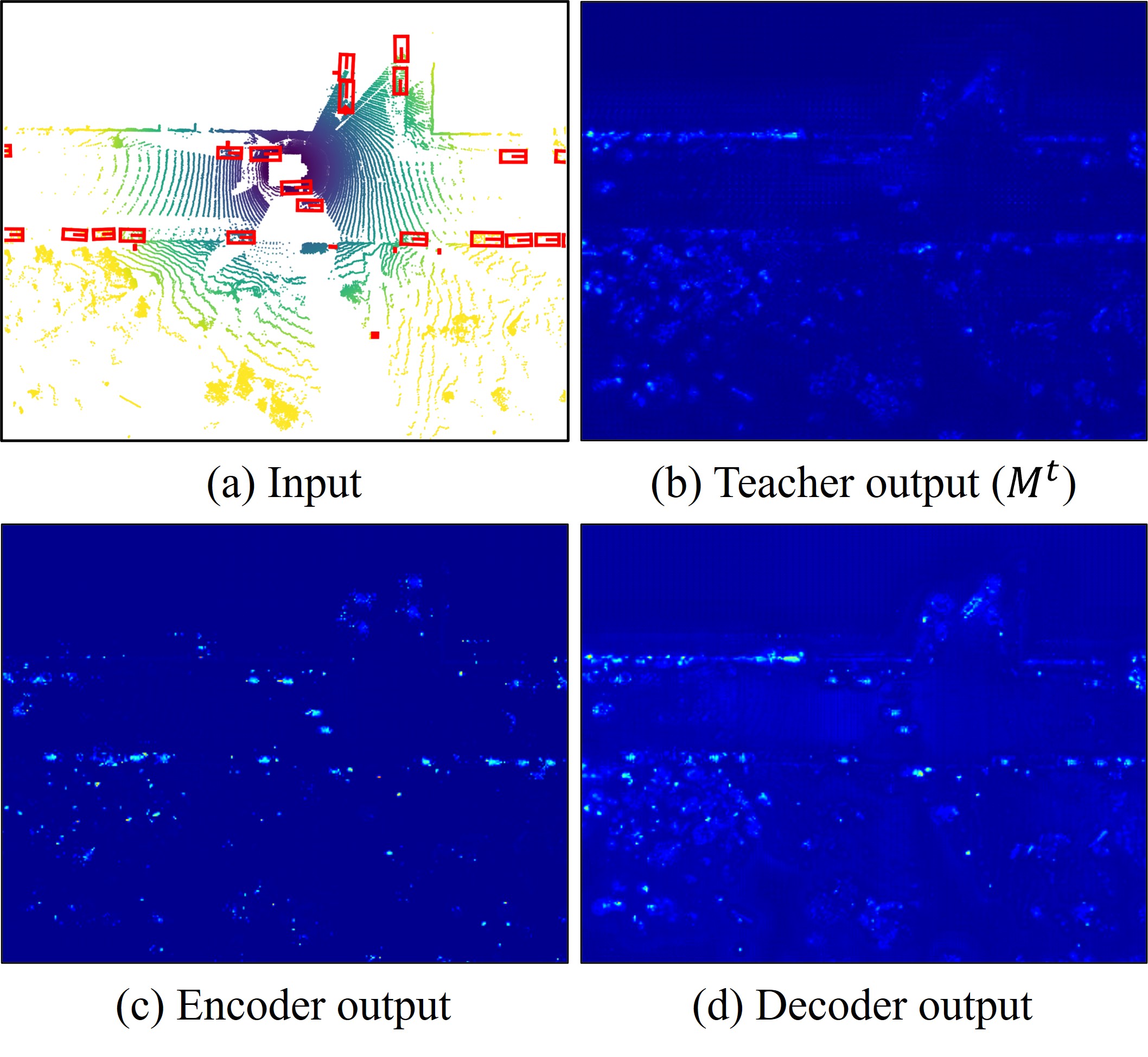}
\vspace{-0.2cm}
\caption{\textbf{Feature visualization on the proposed channel-wise autoencoder.} 
(a) an example input image and (b) the output feature of the teacher network. (c) and (d) are the output images of encoder and decoder of the teacher, respectively. 
}
\vspace{-0.55cm}
\label{fig:fig4}
\end{figure}

As shown in Fig.~\ref{fig:fig4}, the objects and backgrounds are well activated in the example image of the teacher output. On the other hand, the encoder output is activated by further highlighting the coarse positions of the target objects. When looking at the decoder output, we can see that all the fine surrounding information is represented again. At this point, it is worth noting that compared to the teacher output, the target objects are highlighted a little more. From these visual comparisons, we can infer how our method successfully transfers the object-centered knowledge to the student. 

\begin{table}[t]
    \begin{center}
    \caption{\textbf{Buffer layer for different channel size.}} \label{abl:channel}
    \vspace{-3mm}
    \begin{tabular}{c|ccc|c}
    \hline
    Method      & Vehicle        & Pedestrian      & Cyclist & Avg. \\ 
    \hline\hline
    S → T       & 58.41          & \textbf{48.90}  & \textbf{51.90} & \textbf{53.07}   \\
    T → S       & \textbf{58.62} & 48.78           & 51.75 & 53.05  \\
    (S + T) / 2 & 58.47          & 48.84           & 51.54 & 52.95  \\ 
    \hline
    \end{tabular}
    \end{center}
    \vspace{-0.5cm}
\end{table}

We explore the buffer layer that matches the channel size of the channel-wise autoencoder without the head attention loss. As shown in Table~\ref{abl:channel}, we compare the three types for the buffer layer: (1) \textit{S → T} is the upsampling method that increases the student's map-view feature to the teacher's feature. (2) \textit{T → S} is the downsampling method that decreases the teacher's feature to the student's feature. (3) \textit{(S + T) / 2} is that the teacher's feature is downsampled and the student's feature is upsampled to the median size. The experiments show that the upsampling method performs better when considering all the classes.

In Table~\ref{abl:shared}, we observe the performance difference when the autoencoder parameters are shared or not. From the result, we can conclude that the shared parameters achieve better performance because what we want to is for the student to learn the teacher's knowledge, not the independent model.

\begin{table}[t]
    \begin{center}
    \caption{\textbf{Effect of shared and non-shared parameters for the autoencoder.}}   \label{abl:shared}
    \vspace{-3mm}
    \begin{tabular}{c|ccc|c}
    \hline
    Method     & Vehicle & Pedestrian & Cyclist & Avg. \\ \hline\hline
    Non-shared & 56.26   & 45.85      & 48.23 & 50.11   \\
    Shared     & \textbf{58.41}   & \textbf{48.90}      & \textbf{51.90} & \textbf{53.07}   \\ \hline
    \end{tabular}
    \end{center}
    \vspace{-0.5cm}
\end{table}

We investigate improvements made by our interchange transfer for KD without the head attention loss as shown in Table~\ref{tab:recon}. Self-reconstruction is a method wherein the decoder uses the corresponding input for the reconstruction and our interchange reconstruction is a method wherein the proposed $\mathcal{L}_{it}$ objective transfers the reconstructed knowledge to the opponent network. Our interchange transfer-based reconstruction achieves better results and note that our main task is not the reconstruction but the 3D object-based knowledge transfer for KD. 

\begin{table}[t]
    \centering
    \caption{\textbf{Comparison of different reconstruction methods for the autoencoder.}}
    \label{tab:recon}
    \vspace{-3mm}
    \resizebox{\columnwidth}{!}{\begin{tabular}{c|ccc|c}
    \hline
    Method                                                       & Vehicle        & Pedestrian     & Cyclist & Avg.       \\ \hline\hline
    Self Recon.                                                   & 56.57          & 47.26          & 50.29 & 51.37         \\
   Ours & \textbf{58.41} & \textbf{48.90} & \textbf{51.90} & \textbf{53.07} \\ \hline
    \end{tabular}}
    \vspace{-0.2cm}
\end{table}

3D detection~\cite{dai2021dynamic}\cite{ge2020afdet}\cite{CPoint}\cite{rukhovich2022imvoxelnet} has the multiple detection head. To prove the superiority of the proposed head attention objective for 3D object detection, we make the KD comparison results against only multiple detection head without the autoencoder, as shown in Table~\ref{abl:attn}.
Since the heatmap head classifies objects and other heads regress 3D bounding box information, Applying KL loss and $l_1$ loss to all detection heads has a negative effect. However, it is required to consider the relation of detection heads. In this respect, our method achieves better performance than the other KD methods which directly mimic the output of detection heads or simply employ attention mechanism.

\begin{table}[t]
    \centering
    \caption{\textbf{Comparison of KD methods for the multiple detection head.} KL loss and $l_1$ loss denote that directly apply the loss function to all detection heads for KD.}
    \label{abl:attn}
    \vspace{-3mm}
    \resizebox{\columnwidth}{!}{\begin{tabular}{c|ccc|c}
    \hline
    Method                 & Vehicle & Pedestrian & Cyclist & Avg. \\ \hline\hline
    Student                & 55.62   & 46.78      & 49.75  & 50.72 \\ \hline
    Baseline               & 56.26   & 46.75      & 50.48 & 51.16  \\
    KL loss~\cite{hintonKD}& 55.92   & 45.08      & 47.49 & 49.50   \\
    $l_1$ loss             & 55.62   & 45.10      & 48.73  & 49.82 \\ \hdashline
    AT~\cite{AttentionTransfer}             & 56.85   & 47.34      & 50.36  & 51.52 \\
    $\mathcal{L}_{inter}$             & 56.41   & 46.90      & 50.90  & 51.40 \\
    $\mathcal{L}_{intra}$             & 57.20   & 47.19      & 51.23  & 51.87 \\
    $\mathcal{L}_{attn}$   & \textbf{57.10}   & \textbf{47.34}      & \textbf{51.79} & \textbf{52.08}  \\ \hline
    \end{tabular}}
    \vspace{-0.6cm}
\end{table}

Table~\ref{abl:loss} shows the overall effect of the proposed losses on the KD performances. We set up the experiments by adding each loss based on the supervised loss $\mathcal{L}_{sup}$. Specifically, the interchange transfer loss $\mathcal{L}_{it}$ improves on an average of 1.41\% mAPH and the compressed representation loss $\mathcal{L}_{cr}$ leads to a 0.94\% performance improvement. In the end, the head attention loss $\mathcal{L}_{attn}$ helps to improve the performance and the final average mAPH is 53.54\%. We conclude that each proposed loss contributes positively to performance improvement in the 3D object detection-based KD task.

\begin{table}[t]
    \centering
    \caption{\textbf{Ablation results from investigating effects of different components.}} \label{abl:loss}
    \vspace{-3mm}
    \resizebox{\columnwidth}{!}{\begin{tabular}{cccc|ccc|c}
    \hline
    $\mathcal{L}_{sup}$       & $\mathcal{L}_{it}$        & $\mathcal{L}_{cr}$       & $\mathcal{L}_{attn}$      & Vehicle & Pedestrian & Cyclist & Avg. \\ \hline\hline
    
     \checkmark &                           &                           &                           & 55.62   & 46.78      & 49.75  & 50.72 \\
    \checkmark & \checkmark &                           &                           & 57.41   & 48.20      & 50.77  & 52.13 \\
    \checkmark & \checkmark & \checkmark &                           & 58.41   & 48.90      & 51.90   & 53.07\\
    \checkmark & \checkmark & \checkmark & \checkmark & \textbf{58.81}   & \textbf{49.61}      & \textbf{52.19} & \textbf{53.54}  \\ \hline
    \end{tabular}}
    \vspace{-0.2cm}
\end{table}

From Table~\ref{abl:com-det}, we observed quantitative comparisons of the computational complexity between the student network and the teacher network. Specifically, the student network, which reduced the channel by 1/4, decreased about 8.6 times compared to the parameters of the teacher, and FLOPS was reduced by 7.4 times. Above all, we should not overlook the fact that the performance of the student improved from 50.72\% to 53.54\% mAPH/L2 by our KD method.
Furthermore, we apply our method to the student whose channel was reduced by half.
The student's performance increases to 59.04\%, and the parameters and FLOPS compared to the teacher are reduced by 3.5 times and 2.6 times, respectively. Compared to lightweight network-based methods~\cite{PP}\cite{SECOND}\cite{shi2020points}\cite{zhang2022not}, our student networks are able to derive stable performance with fewer parameters and FLOPS in 3D object detection.

\begin{table}[t]
    \begin{center}
    \caption{\textbf{Quantitative evaluation for model efficiency on Waymo dataset.}}
    \label{abl:com-det}
    \vspace{-3mm}
    \resizebox{\columnwidth}{!}{\begin{tabular}{c|ccc}
    \hline
    Method                            & Params (M)  & FLOPS (G) & mAPH / L2 \\ \hline\hline
    PointPillars~\cite{PP}            & 4.8         & 255.0     & 57.05 \\
    SECOND~\cite{SECOND}              & 5.3         & 84.5      & 57.23 \\
    Part-A$^2$~\cite{shi2020points}   & 4.6         & 87.1      & 57.43 \\
    IA-SSD~\cite{zhang2022not}        & 2.7         & 46.1      & 58.08 \\
    SparseKD-v0.64~\cite{yang2022towards} & 5.2 & 85.1 & 58.89\\
    \hline
    Teacher~\cite{CPoint}             & 5.2         & 333.9     & 60.13 \\
    Ours: Student (${1}/{2}$)         & 1.5         & 130.1     & 59.04 \\
    Ours: Student (${1}/{4}$)         & 0.6         & 45.1      & 53.54 \\\hline
    \end{tabular}}
    \end{center}
    \vspace{-0.7cm}
\end{table}

\section{Conclusion}
In this paper, we propose a novel KD method that transfers knowledge to produce a lightweight point cloud detector. Our main method involves interchange transfer, which learns coarse knowledge by increasing the similarity of the compressed feature and fine knowledge by decompressing the map-view feature of the other side using the channel-wise autoencoder.
Moreover, we introduce a method to guide multiple detection head using head relation-aware self-attention, which refines knowledge by considering the relation of instances and properties.
Ablation studies demonstrate the effectiveness of our proposed algorithm, and extensive experiments on the two large-scale open datasets verify that our proposed method achieves competitive performance against state-of-the-art methods.

\noindent\small\textbf{Acknowledgement.}
This work was partly supported by NRF-2022R1A2C1091402, BK21 FOUR program of the NRF of Korea funded by the Ministry of Education (NRF5199991014091), and IITP grant funded by the Korea government(MSIT) (No.2021-0-00951, Development of Cloud based Autonomous Driving AI learning Software; No. 2021-0-02068, Artificial Intelligence Innovation Hub). W. Hwang is the corresponding author. 

{\small
\bibliographystyle{ieee_fullname}
\bibliography{main}
}

\newpage




\section{Student configuration}
\begin{table}[ht]
\centering
\caption{The number of parameters of the teacher, the student (${1}/{2}$), and the student (${1}/{4}$).}
\label{tab:params}
\resizebox{\columnwidth}{!}{
\begin{tabular}{c|cccc}
\hline
Model               & Point cloud encoder & Backbone & Head   & Total          \\ \hline\hline
Teacher             & 4,608                & 4,806,400  & 413,003 & 5,224,011 (5.2M) \\
Student (${1}/{2}$) & 4,608                & 1,212,288  & 302,411 & 1,519,307 (1.5M) \\
Student (${1}/{4}$) & 4,608                & 308,416   & 247,115 & 560,139 (0.6M)  \\ \hline
\end{tabular}
}
\end{table}
Conventional KD methods for 3D object detection focused on improving performance or reducing latency. However, the main purpose of our method is how to reduce the parameters of 3D object detector.
In this respect, we investigate which component of the backbone architecture has most parameters as shown in Table~\ref{tab:params}. Since the backbone has 4.8M parameters, which occupies about 92\% of the 5.2M parameters of the teacher network, We apply channel reduction to each layers of backbone because channel reduction maintains performance better than depth reduction on detection task~\cite{MimicKD}\cite{wang2019distilling}. Finally, our student (${1}/{4}$) has 8.7$\times$ less parameters and student (${1}/{2}$) has 3.5$\times$ less parameters.

\section{Performance of the student ${1}/{2}$}
\begin{table}[h]
\centering
\caption{Comparison with different KD methods in Waymo and nuScenes validation set.}
\label{tab:comp-kd}
\resizebox{\columnwidth}{!}{%
\begin{tabular}{c|cccc|cc}
\hline
\multirow{2}{*}{Method} & \multicolumn{4}{c|}{Waymo}                                                           & \multicolumn{2}{c}{nuScenes} \\
                        & Vehicle             & Pedestrian          & Cyclist              & Total             & NDS                    & mAP          \\\hline\hline
Teacher~\cite{CPoint}        & 65.11               & 54.99               & 60.28                & 60.13             & 59.45                  & 48.83        \\
Student                 & 62.19               & 53.19               & 56.45                & 57.28             & 56.79                  & 45.45        \\\hline
Baseline                & 63.16               & 53.81               & 57.21                & 58.06             & 57.95                  & 46.78        \\
FitNet~\cite{FitNet}                  & \underline{63.45}   & 54.10               & 57.31                & 58.29             & 57.97                  & 46.78        \\
EOD-KD~\cite{detKD}                 & 62.80               & 53.70               & 57.27                & 57.92             & \underline{58.07}                  & 46.83        \\
TOFD~\cite{TOFD}                    & 60.99               & 52.98               & 57.44                & 57.14             & 57.54                  & 46.10        \\
SE-SSD~\cite{zheng2021se}                  & 63.02               & 54.21   & 57.86                & \underline{58.36} & 57.30                  & 46.03        \\
Obj. DGCNN~\cite{objDGCNN}              &  63.07              & \underline{54.23}            & 57.77                &  \underline{58.36}            & 57.96                  & \underline{46.92}        \\
SparseKD~\cite{yang2022towards}                & 62.57               & 53.75               & \underline{58.06}    & 58.13             &  57.59                 &   46.54      \\\hline
Ours                    & \textbf{63.92}      & \textbf{54.53}      & \textbf{58.66}       & \textbf{59.04}    & \textbf{58.32}         & \textbf{47.18}        \\
\hline
\end{tabular}%
}
\end{table}
To verify the generality of our method, we compare the student (${1}/{2}$) with other KD methods on Waymo and nuScenes validation set. Table~\ref{tab:comp-kd} shows the mAPH of level2 performance of KD methods on Waymo, and NDS and mAP on nuScenes. Our student (${1}/{2}$) shows better performance than other methods on both datasets. In conclusion, we confirm that our method has generality regardless of the parameter reduction ratio.

\section{Pseudocode}
\begin{algorithm}[ht]
\scriptsize
\SetAlgoLined
    \PyDef{c\_t: Channel size of the teacher's backbone output} \\
    \PyDef{c\_s: Channel size of the student's backbone output} \\
    \PyDef{c\_e: Channel size of the compressed representation} \\
    \PyDef{x\_t: The map-view feature of the teacher network} \\
    \PyDef{x\_s: The map-view feature of the student network} \\
    \PyNline{} \\
    \PyComment{Define the channel-wise autoencoder as class} \\
    \PyClass{class }\PyCode{ChannelWiseAE(nn.Module):} \\
    \Indp   
        \PyClass{def }\PyCode{\_\_init\_\_(self, c\_t, c\_s, c\_e):} \\
        \Indp   
            \PyComment{Sampling layers to adapt channel size} \\
            \PyCode{self.downs = 
            nn.Conv2d(c\_t, c\_s, (1, 1))}\\
            \PyCode{self.ups = 
            nn.Conv2d(c\_s, c\_t, (1, 1))} \\
            \PyComment{Build encoder layers} \\
            \PyCode{self.encoder = \\
            \Indp
            nn.Sequential(\\
            \Indp
            nn.Conv2d(c\_t, 128, (1, 1)), \\
            nn.Conv2d(128, 64, (1, 1)), \\
            nn.Conv2d(64, c\_e, (1, 1))) \\
            \Indm
            \Indm
            }
            
            \PyComment{Build decoder layers} \\
            \PyCode{self.decoder = \\
            \Indp
            nn.Sequential(\\
            \Indp
            nn.Conv2d(c\_e, 64, (1, 1)), \\
            nn.Conv2d(64, 128, (1, 1)), \\
            nn.Conv2d(128, c\_t, (1, 1))) \\
            \Indm
            \Indm
            }
            
        \Indm 
        \PyClass{def }\PyCode{forward(self, x\_t, x\_s):} \\
            \Indp
            \PyComment{Pass through the autoencoder} \\
            \PyCode{
                x\_s = self.ups(x\_s) \\
                comp\_t = self.encoder(x\_t) \\
                comp\_s = self.encoder(x\_s) \\
                decomp\_t = self.decoder(comp\_t) \\
                decomp\_s = self.decoder(comp\_s) \\
                decomp\_t = self.downs(decomp\_t) \\
            }
            \PyComment{Calculate loss values} \\
            \PyCode{
                comp\_repr = F.l1\_loss(comp\_s, comp\_t) \\
                decomp\_s2t = F.l1\_loss(s\_decode, x\_t) \\
                decomp\_t2s = F.l1\_loss(t\_decode, x\_s) \\
            }
            \PyComment{Return total loss} \\
            \PyClass{return }\PyCode{comp\_repr + decomp\_s2t + decomp\_t2s} \\
            \Indm
    \Indm 
\caption{\small PyTorch-style pseudocode for the channel-wise autoencoder}
\label{algo:autoencoder}
\end{algorithm}
\begin{algorithm}[ht]
\scriptsize
\SetAlgoLined
    \PyDef{x\_t: Detection results of the teacher network} \\
    \PyDef{x\_s: Detection results of the student network} \\
    \PyDef{fusion: 1$\times$1 convolution layer for fusion on channel dimension} \\
    \PyDef{ind: Index of objects' location} \\
    \PyNline{} \\
    \PyComment{Define the self-attention} \\
    \PyClass{def }\PyCode{self\_attention(x):} \\
        \Indp   
            \PyComment{Calculate attention score} \\
            \PyCode{score = F.softmax(torch.matmul(x.transpose(-2, -1), x) / torch.sqrt(x.size(-2)), dim=-2)} \\
            \PyClass{return }\PyCode{torch.matmul(x, score)} \\
        \Indm
    \PyNline{} \\
    \PyComment{Define the head relation-aware self-attention} \\
    \PyClass{def }\PyCode{relation\_aware\_self\_attention(x):} \\
        \Indp   
        \PyComment{Generate feature sequences} \\
        seq = x.gather(ind) \\
        \PyComment{Apply the intra-head relation attention} \\
        \PyClass{for }\PyCode{seq\_head in seq:} \\
        \Indp
            intra\_at1tention.append(self\_attention(seq\_head)) \\
        \Indm
        intra\_attention = torch.cat(inter\_attention, dim=1) \\
        \PyComment{Apply the inter-head relation attention} \\
        inter\_attention = (self\_attention(seq)) \\
        \PyComment{Pass through the fusion layer} \\
        attention = \\
        \Indp
            fusion(torch.cat([intra\_attention, inter\_attention], dim=1)) \\
        \Indm
        
        \PyClass{return }\PyCode{attention} \\
        \Indm
    \PyNline{} \\
    \PyComment{Apply the relation-aware self-attention} \\
    \PyCode{
        rasa\_t = relation\_aware\_self\_attention(x\_t) \\
        rasa\_s = relation\_aware\_self\_attention(x\_s) \\
        \PyComment{Calculate the attentive head loss} \\
        attentive\_head = F.l1\_loss(rasa\_s, rasa\_t) \\
    }
    \PyComment{Return the loss} \\
    \PyClass{return }\PyCode{attentive\_head} \\
\caption{\small PyTorch-style pseudocode for the relation-aware self-attention}
\label{algo:attention}
\end{algorithm}

Algorithm~\ref{algo:autoencoder} and \ref{algo:attention} show PyTorch-style pseudo-code for the channel-wise autoencoder and the head relation-aware self-attention, respectively.
The interchange transfer and the compressed representation loss are included in Algorithm~\ref{algo:autoencoder}. Algorithm~\ref{algo:attention} contains the head attention loss. As we described in section \textcolor{red}{3.3}, we use the $l_1$ loss as a similarity function. 

\begin{figure}[ht]
\centering
\includegraphics[width=1\linewidth]{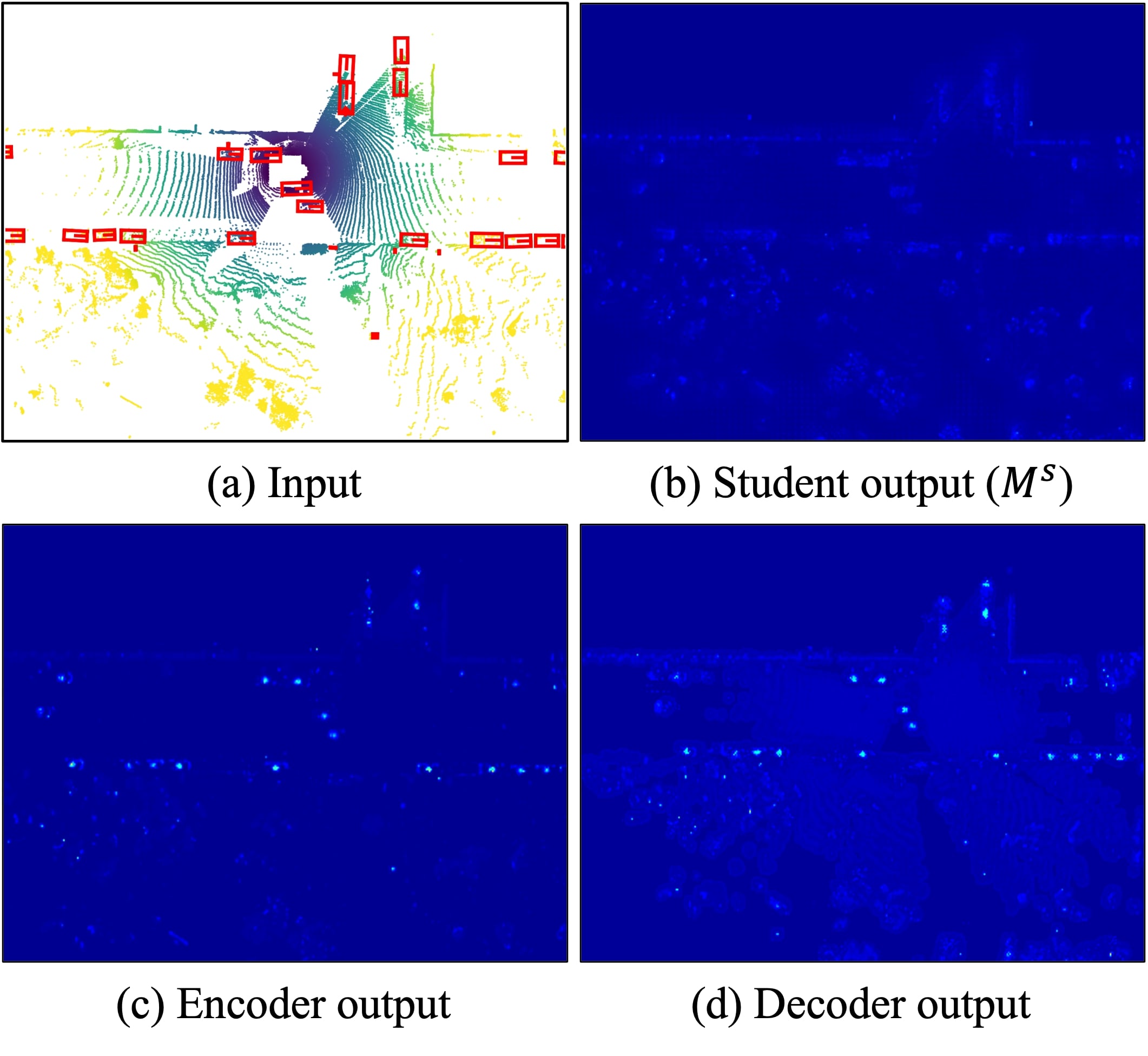}
\vspace{-0.5cm}
\caption{\textbf{Output feature visualization of the student backbone.}}
\label{fig:vis_student}
\end{figure}

\section{Visualization of the student feature} 
We visualize the output features of the student, the encoder, and the decoder, which take the same input as in Fig.~\ref{fig:fig4} of the main paper. As shown in Fig.~\ref{fig:vis_student}, the visualization results show that both objects and backgrounds are well-activated.

\section{Inference time} 
\begin{wraptable}{l}{0.5\linewidth}
\caption{Lantency and FPS.}
\vspace{-0.3cm}
\label{tab:inference}
\small
\centering
\begin{tabular}{c|cc}
\hline
Model   & latency & FPS  \\ \hline\hline
Teacher & 46.0    & 21.7 \\
Ours & 23.4    & 42.7 \\ \hline
\end{tabular}
\end{wraptable}
Table~\ref{tab:inference} shows the inference time of the teacher and our student (Ours, ${1}/{4}$). The inference time is averaged 100 frames with a NVIDIA Titan V.  Our student network achieves a computation speed of 42.7 FPS.

\section{Performance of voxel-based encoders} 

\begin{table}[h]
\caption{Performances on the voxel-based encoder.}
\label{tab:voxel}
\vspace{-0.3cm}
\resizebox{\columnwidth}{!}{
\begin{tabular}{c|ccccc}
\hline
Method & Teacher & Student & Baseline & SparseKD    & Ours           \\ \hline\hline
mAPH/L2 & 65.50   & 63.26   & 64.03    & 64.05 & \textbf{64.26} \\ \hline
\end{tabular}}
\vspace{-0.4cm}
\end{table}

We made additional experiments in Table~\ref{tab:voxel} that shows the results of the voxel-based encoder. Our method shows 64.26\% mAPH/L2 and outperforms SparseKD, which is the latest KD method for 3D object detectors.

\section{Limitation}
The limitation of the interchange transfer lies in the fact that both the teacher and the student networks must maintain the same spatial resolution, as the interchange transfer is based on feature-based knowledge distillation.
We also note that using the autoencoder often requires additional effort for identifying the proper network structure or its hyper-parameters for the different 3D object detection, but we believe that the deviations of the optimal hyper-parameters are not high. 

\section{Potential negative societal impacts}
Our KD method aims to make an efficient 3D object detection network, which is crucial for the autonomous driving system that requires real-time response. One potential negative societal impact of our method is that the quantitative performance of the student network follows similarly to that of the teacher network; also, it has not been confirmed whether there are any parts that can be fatal to the safety of the autonomous driving system in the wild.

\end{document}